\documentclass[11pt,twocolumn]{article}

\usepackage[utf8]{inputenc}
\usepackage[T1]{fontenc}
\usepackage{times}
\usepackage[margin=0.75in,columnsep=0.3in]{geometry}
\usepackage{amsmath,amssymb}
\usepackage{graphicx}
\usepackage{booktabs}
\usepackage{array}
\usepackage{hyperref}
\usepackage{xcolor}
\usepackage{listings}
\usepackage{enumitem}
\usepackage{caption}
\usepackage{float}
\usepackage{fancyhdr}
\usepackage{titlesec}
\usepackage{wasysym}
\usepackage{tikz}

\definecolor{codegreen}{rgb}{0.25,0.49,0.48}
\definecolor{codegray}{rgb}{0.5,0.5,0.5}
\definecolor{codebg}{rgb}{0.97,0.97,0.97}
\definecolor{linkblue}{rgb}{0.0,0.33,0.71}

\hypersetup{
  colorlinks=true,
  linkcolor=linkblue,
  citecolor=linkblue,
  urlcolor=linkblue,
}

\titleformat{\section}{\large\bfseries}{\thesection}{1em}{}
\titleformat{\subsection}{\normalsize\bfseries}{\thesubsection}{1em}{}
\titleformat{\subsubsection}{\normalsize\itshape}{\thesubsubsection}{1em}{}
\titlespacing*{\section}{0pt}{12pt}{6pt}
\titlespacing*{\subsection}{0pt}{8pt}{4pt}

\setlist{nosep,leftmargin=1.5em}

\begin{document}

\twocolumn[
\begin{@twocolumnfalse}
\begin{center}
{\LARGE\bfseries Reasoning Provenance for Autonomous AI Agents:\\[4pt]
Structured Behavioral Analytics Beyond\\State Checkpoints and Execution Traces\par}
\vspace{12pt}
{\large Neelmani Vispute \qquad Aditya Kadam\par}
\vspace{4pt}
{\normalsize Oracle Cloud Infrastructure\par}
\vspace{4pt}
{\small\texttt{\{neelmani.vispute, aditya.kadam\}@oracle.com}\par}
\vspace{10pt}
{\small March 2026\par}
\vspace{16pt}
\end{center}

\begin{center}
\parbox{0.92\textwidth}{
\small
\textbf{Abstract.}
As AI agents transition from human-supervised copilots to autonomous platform infrastructure, the ability to analyze their reasoning behavior across populations of investigations---not just debug individual executions---becomes a pressing infrastructure requirement. Existing operational tooling addresses adjacent needs effectively: state checkpoint systems (LangGraph) enable fault tolerance; observability platforms (LangSmith, Langfuse, Datadog) provide execution traces for debugging; telemetry standards (OpenTelemetry GenAI) ensure interoperability; provenance models (PROV-AGENT) capture causal lineage for scientific reproducibility. These systems, individually and in combination, provide strong coverage of the \emph{mechanical layer} of agent execution.

What current systems do not natively provide as a first-class, schema-level primitive is \emph{structured reasoning provenance}---normalized, queryable records of why the agent chose each action, what it concluded from each observation, how each conclusion shaped its strategy, and which evidence supports its final verdict. While reasoning-like information may exist within trace data as unstructured natural language in model responses, or may be partially approximated through extensible metadata fields, it is not captured as a canonical, comparable semantic object designed for cross-run behavioral analytics.

This paper introduces the \textbf{Agent Execution Record (AER)}, a structured reasoning provenance primitive that captures intent, observation, and inference as first-class queryable fields on every step, alongside versioned plans with revision rationale, evidence chains, structured verdicts with confidence scores, and delegation authority chains. We formalize the distinction between computational state persistence and reasoning provenance, argue that the latter cannot in general be faithfully reconstructed from the former, and show how AERs enable population-level behavioral analytics: reasoning pattern mining, confidence calibration, cross-agent comparison, and counterfactual regression testing via mock replay.

We present a domain-agnostic model with extensible domain profiles, a reference implementation and SDK, and outline an evaluation methodology informed by preliminary deployment on a production platformized root cause analysis agent. We further introduce a transport-layer verification mechanism---independent interception of tool calls at the stdio, shell, and filesystem layers---that enables post-hoc reconciliation of self-reported AER fields against ground-truth execution data, producing per-investigation fidelity scores. Preliminary storage analysis suggests AERs are substantially more compact than cumulative state checkpoints while retaining higher long-term analytical value. The paper's primary contribution is the AER abstraction and its formalization; empirical validation across diverse workloads is ongoing work.

\vspace{6pt}
}
\end{center}
\vspace{16pt}
\end{@twocolumnfalse}
]

\section{Introduction}

The operational model for AI agents is shifting from interactive copilots to fully autonomous platform infrastructure. In the copilot model, the human operator serves as the implicit audit trail---they observe the agent's reasoning, validate outputs, and can reconstruct decisions from memory. In the autonomous model, agents are triggered by events, execute investigation plans in seconds without oversight, and produce outputs that feed downstream systems or other agents. The only record of the agent's reasoning is what the infrastructure captures.

\paragraph{Running example.} Throughout this paper, we use a concrete incident to illustrate AER's design. A platformized root cause analysis (RCA) agent, serving multiple engineering teams in a multi-cloud infrastructure organization, is triggered by Jira ticket DBINFRA-1458: ``DB connectivity failures on SCAN IP.'' The agent operates fully autonomously---it reads the ticket, retrieves relevant runbook context, formulates an investigation plan, executes it by querying monitoring systems and database tools, encounters unexpected evidence that causes it to revise its plan, and ultimately diagnoses the root cause as an out-of-memory (OOM) condition on a RAC cluster node. This single investigation illustrates every component of the AER model.

Current systems offer two well-developed answers to the question of what to capture. \textbf{State checkpoint systems} (exemplified by LangGraph's checkpointer) save serialized snapshots of the agent's computational state at every execution step, enabling fault tolerance, time-travel debugging, and human-in-the-loop workflows. \textbf{Observability platforms} (LangSmith, Langfuse, Datadog) capture execution traces showing every tool call, LLM interaction, token usage, and latency. Together---and they are often used together---these provide comprehensive operational tooling for individual agent executions.

But as organizations scale to platformized agents handling hundreds of investigations daily, a different class of question emerges:

\begin{itemize}
\item \emph{``In what percentage of database incidents does the agent re-plan after the first step?''}
\item \emph{``When our agent reports confidence below 0.7, how often does a human expert disagree with the verdict?''}
\item \emph{``Would the new prompt version produce systematically different reasoning across 200 historical incidents?''}
\item \emph{``What evidence pattern most commonly precedes a wrong diagnosis?''}
\end{itemize}

These are \textbf{population-level behavioral analytics questions}. They require structured, queryable reasoning data across thousands of investigations. Answering them from individual traces or serialized state snapshots is not impossible in principle---one could build custom extraction pipelines, attach metadata to traces, or run NLP over model responses. But these are \emph{post-hoc reconstruction} approaches operating on data not designed for this purpose. The resulting representations are fragile, unstandardized across runs, and lack the schema guarantees that enable reliable cross-run comparison.

This paper introduces the \textbf{Agent Execution Record (AER)}, an infrastructure primitive that captures \emph{reasoning provenance}---structured annotations of agent intent, observation, and inference---at execution time, as a first-class schema-level construct. AER operates at a different analytical level from state checkpoints and execution traces. The relationship is analogous to the distinction between Application Performance Monitoring (APM) and Business Intelligence (BI): APM tells you a request took 340ms and called three services; BI tells you conversion dropped 12\% in the southeast after a pricing change. Both valuable, both necessary, operating at different analytical levels. AERs provide the BI layer for autonomous agent reasoning.

The contributions of this paper are:

\begin{itemize}
\item \textbf{A formalized distinction} between computational state persistence and reasoning provenance (Section~3), with a non-identifiability argument showing that reasoning provenance cannot in general be faithfully reconstructed as a stable, queryable representation from state checkpoints alone.
\item \textbf{The AER model} (Section~4), capturing intent, observation, and inference as first-class queryable fields, alongside versioned plans, evidence chains, structured verdicts, delegation authority chains, and retrieval provenance---with extensible domain profiles. Illustrated throughout with the DBINFRA-1458 running example.
\item \textbf{Three replay modes} (Section~5), with mock replay enabling counterfactual model/prompt comparison on real production data without live system access---a capability not natively provided by existing systems.
\item \textbf{A reference implementation and preliminary analysis} (Sections~6--7), including a file-based SDK, a streaming scale-out architecture, preliminary storage comparison with cumulative checkpoints, and an evaluation methodology for the behavioral analytics claims.
\end{itemize}

\section{Related Work}

We survey existing approaches not to argue they are inadequate---they are excellent at their intended purposes---but to precisely locate the analytical level at which AER operates. Our claim is not that existing systems \emph{cannot} capture reasoning information; it is that they do not \emph{natively represent} it as a canonical, comparable, schema-constrained semantic object designed for cross-run behavioral analytics.

\subsection{State Checkpoint Systems (LangGraph)}

LangGraph's checkpointer saves serialized \emph{MessagesState} snapshots (message history, channel values, tool call results) at every super-step, persisted to PostgreSQL, DynamoDB, or SQLite. It enables time-travel debugging, human-in-the-loop workflows, conversational memory, and fault tolerance. This is mature, production-grade infrastructure that AER does not aim to replace.

A checkpoint for step 2 of the DBINFRA-1458 investigation would contain the accumulated message history including the user's initial request, the agent's prior responses, and the tool call result ``Node3: TNS-12541 no listener.'' A human reading this checkpoint \emph{can} infer what the agent was doing. What the checkpoint does not contain as a structured field is \emph{why} the agent chose to check the listener at this point (intent), that the agent \emph{concluded} this directly explains the reported connectivity failures (observation), or that this conclusion caused the agent to \emph{abandon its original plan and formulate a new one} (inference + plan revision). This information may exist as unstructured text within the model's response, but it is not a normalized, queryable field.

Checkpoints are also \emph{cumulative}---each contains all prior state. Preliminary analysis (Section~7) suggests this results in substantially higher per-investigation storage than AER, with value that decays rapidly after execution completes, whereas AER value appreciates as investigations become regression test cases and pattern references.

\subsection{Observability Platforms (LangSmith, Langfuse, Datadog)}

Observability platforms capture execution trees with tool calls, LLM inputs/outputs, tokens, and latency. LangSmith supports Annotation Queues for expert review, dataset-based evaluation for regression testing, and monitoring dashboards. These capabilities are significant: combined with a checkpointer, an observability platform provides comprehensive operational visibility into individual agent runs.

We acknowledge that observability platforms are more extensible than early versions of this work suggested. Custom spans, metadata tags, evaluation scores, and downstream analytics pipelines \emph{can} be built on top of traces. However, the resulting reasoning representations are \emph{ad-hoc}: they depend on implementation-specific conventions, lack cross-run schema guarantees, and do not form a canonical data model designed for population-level behavioral comparison. The difference is analogous to storing customer addresses in free-text notes versus structured city/state/zip columns.

\subsection{OTel GenAI, Provenance Models, and Audit Trails}

OpenTelemetry GenAI semantic conventions standardize telemetry transport (model spans, agent spans, tool execution)---an important interoperability layer that AER can emit as a side effect. PROV-AGENT~\cite{souza2025} extends W3C PROV with agent-centric entities for scientific workflow provenance, providing formal causal lineage. Agent audit trail systems capture identity, authorization, and delegation chains for compliance. Each addresses a real need; AER subsumes audit trail capabilities via its delegation chain while adding the reasoning layer these systems were not designed to capture.

\subsection{Positioning}

Table~\ref{tab:positioning} summarizes how each approach addresses different analytical levels. The symbols indicate: $\bigstar$ = native, first-class support; $\LEFTcircle$ = achievable through extension or custom work; $\circ$ = not a design target.

\begin{table*}[t]
\centering
\small
\caption{Analytical levels and support across existing approaches.}
\label{tab:positioning}
\begin{tabular}{@{}l c c c c c c@{}}
\toprule
\textbf{Analytical Level} & \textbf{LG Chkpt} & \textbf{Obs.\ Plat.} & \textbf{OTel} & \textbf{Prov.} & \textbf{Audit} & \textbf{AER} \\
\midrule
Fault tolerance / resumption       & $\bigstar$ & $\circ$ & $\circ$ & $\circ$ & $\circ$ & $\circ$ \\
Per-run debugging / tracing        & $\LEFTcircle$ & $\bigstar$ & $\bigstar$ & $\LEFTcircle$ & $\circ$ & $\LEFTcircle$ \\
Identity / authority provenance    & $\circ$ & $\circ$ & $\circ$ & $\LEFTcircle$ & $\bigstar$ & $\bigstar$ \\
Structured reasoning provenance    & $\circ$ & $\LEFTcircle$ & $\circ$ & $\LEFTcircle$ & $\circ$ & $\bigstar$ \\
Population-level behavioral analytics & $\circ$ & $\LEFTcircle$ & $\circ$ & $\circ$ & $\circ$ & $\bigstar$ \\
Mock replay / counterfactual testing & $\circ$ & $\circ$ & $\circ$ & $\circ$ & $\circ$ & $\bigstar$ \\
Transport-layer verification       & $\circ$ & $\circ$ & $\circ$ & $\circ$ & $\circ$ & $\bigstar$ \\
\bottomrule
\end{tabular}

\vspace{4pt}
{\scriptsize \emph{Note:} $\LEFTcircle$ for observability platforms acknowledges that custom pipelines can approximate these capabilities. AER's claim is native, schema-level support---not exclusive capability.}
\end{table*}

\section{Computational State vs.\ Reasoning Provenance}

This section formalizes the distinction motivating AER. We define both constructs, state the observability boundary, and argue that reasoning provenance cannot in general be faithfully reconstructed from computational state as a stable, queryable representation.

\subsection{Definitions}

\begin{description}[style=unboxed,leftmargin=0pt]
\item[Definition 1 (Computational State).] The computational state at step $k$ of an agent execution is the persisted snapshot $S_k = (M_k, C_k, T_k)$ where $M_k$ is the accumulated message history, $C_k$ is the set of framework-specific channel values, and $T_k$ is the set of tool call records (function name, arguments, return value). This corresponds to what LangGraph's checkpointer persists at each super-step boundary.

\item[Definition 2 (Reasoning Provenance).] The reasoning provenance at step $k$ is the tuple $R_k = (I_k, O_k, N_k, P_k)$ where $I_k$ is a structured intent statement (why the agent chose this action), $O_k$ is a structured observation (what the agent concluded from the tool result in context), $N_k$ is a structured inference (how this conclusion updates the agent's strategy), and $P_k$ is the plan version that motivated the step. Each component is a normalized, typed field with a stable schema across runs.

\item[Definition 3 (Observability Boundary).] The observability boundary is the set of information available for post-hoc analysis after execution completes. Under state checkpointing, this includes $S_{1..K}$ (all checkpoints). Under combined checkpointing and tracing, it additionally includes the execution tree of spans. Crucially, \emph{intermediate model generations}---the full text of every LLM response---are included within the observability boundary. We do not assume any information is hidden.
\end{description}

\subsection{Non-Identifiability Argument}

We do not claim that reasoning provenance is \emph{logically impossible} to extract from computational state. Our argument concerns \emph{faithful reconstruction as a stable, queryable representation}.

\textbf{Proposition.} Given only the persisted computational state $S_{1..K}$, reasoning provenance $R_{1..K}$ cannot in general be faithfully reconstructed as a normalized, schema-conforming, cross-run-comparable representation without contemporaneous capture at execution time.

\textbf{Argument.} We identify three sources of non-identifiability:

\textbf{(1) Intent multiplicity.} Consider two independent runs of the DBINFRA-1458 investigation. In run A, the agent checks the listener because its plan is to systematically check each infrastructure layer top-down. In run B, the agent checks the listener because it detected ``TNS'' in the error message. Both produce \emph{identical} tool calls and state at step 2. The intent differs. A post-hoc extractor may find a rationale in the model's response, but (a) the model may not state its intent, (b) the stated intent may be post-hoc rationalization, and (c) across runs, extracted text will have different surface forms, preventing schema-normalized comparison.

\textbf{(2) Observation ambiguity.} The tool returns ``Node1: OK, Node2: OK, Node3: TNS-12541 no listener.'' The state records this string. The \emph{observation}---``Node 3 listener is down, which directly explains the client connectivity failures reported in the ticket''---is the agent's contextual interpretation. Extracting this from surrounding text (planning, hedging, formatting) is an NLP extraction problem with inherent error.

\textbf{(3) Inference volatility.} After observing the listener failure, the agent's inference is: ``Proximate cause found. Revising plan to investigate Node 3.'' This forward-looking strategic reasoning may or may not appear in the response, and its form varies with model family, temperature, and prompt style. Across 1,000 investigations, post-hoc-extracted inferences will have high variance in form and completeness.

We therefore distinguish three approaches:
\begin{itemize}
\item \textbf{Post-hoc extraction}: NLP on model responses. Possible but fragile and non-standardized.
\item \textbf{Extensible annotation}: custom metadata on trace spans. Possible if implemented, but ad-hoc.
\item \textbf{Contemporaneous structured capture (AER)}: explicit, schema-constrained fields at execution time. Stable, normalized, cross-run comparable by design.
\end{itemize}

\subsection{Limitation: Self-Reported Reasoning}

A necessary acknowledgment: AER's reasoning fields are populated by the agent itself. They record what the agent \emph{reports} about its reasoning, not a ground-truth account of its internal computation. This creates several known risks:

\begin{itemize}
\item \textbf{Post-hoc rationalization}: the agent may construct a coherent narrative that does not reflect its actual decision process.
\item \textbf{Format gaming}: the agent may produce plausible-sounding but vacuous reasoning annotations.
\item \textbf{Model-family style differences}: different LLM families produce systematically different self-explanation styles.
\item \textbf{Prompt-induced verbosity}: prompts requesting structured reasoning may elicit more words without more insight.
\end{itemize}

We argue that self-reported reasoning, despite these limitations, is (a) substantially more useful for behavioral analytics than no structured reasoning at all; (b) analogous to established practices in medicine and law, where case notes and legal reasoning opinions are self-reported and still foundational for audit and quality improvement; and (c) \emph{empirically validatable} through the very infrastructure AER provides. Specifically, AER's structured verdict with confidence scores enables calibration against ground truth (human expert assessments), and action/intent consistency can be measured across the AER corpus (do agents that report ``rule out DNS'' actually perform DNS-related checks?). The infrastructure for capturing structured reasoning must exist before the faithfulness of that reasoning can be studied systematically.

However, acknowledging the limitation is insufficient. Section~3.4 introduces a transport-layer verification mechanism that provides independent ground truth against which self-reported AER fields can be validated.

\subsection{Transport-Layer Verification}

AER's reasoning fields (intent, observation, inference) are self-reported, but the \emph{tool calls} recorded in steps.jsonl---which tools were invoked, with what inputs, producing what outputs---need not rely on self-report. We introduce a three-layer interception architecture (Figure~\ref{fig:architecture}) that captures ground-truth execution data independently of the agent, enabling post-hoc reconciliation.

\begin{figure*}[t]
\centering
\begin{tikzpicture}[
  box/.style={draw, rounded corners=3pt, minimum height=0.7cm, font=\small, align=center, line width=0.4pt},
  agent/.style={box, fill=purple!8, draw=purple!40},
  aer/.style={box, fill=red!6, draw=red!30, minimum width=2.0cm, font=\scriptsize},
  tool/.style={box, fill=teal!8, draw=teal!30},
  intercept/.style={box, fill=green!8, draw=green!30, minimum width=2.4cm},
  recon/.style={box, fill=blue!8, draw=blue!30, minimum width=5.0cm},
  score/.style={box, fill=orange!10, draw=orange!40, rounded corners=8pt},
  trigger/.style={box, fill=gray!8, draw=gray!30},
  lbl/.style={font=\scriptsize\bfseries},
  sublbl/.style={font=\scriptsize, text=gray},
  arr/.style={->, >=stealth, line width=0.4pt},
  darr/.style={->, >=stealth, line width=0.3pt, dashed, gray},
]

\draw[dashed, gray!40, rounded corners=8pt, line width=0.4pt] (-5.8, -5.4) rectangle (5.8, 0.6);
\node[font=\scriptsize, text=gray!60] at (-3.8, 0.4) {Container (agent cannot modify interceptors or their logs)};

\node[trigger, minimum width=2.8cm] (trig) at (0, 1.4) {Event trigger};

\node[agent, minimum width=2.2cm, minimum height=0.85cm] (agent) at (0, 0) {\textbf{Agent}};
\draw[arr] (trig) -- (agent);

\node[lbl] at (-3.8, -0.8) {Self-reported};
\node[sublbl] at (-3.8, -1.1) {Agent writes};

\draw[arr] (agent.south west) -- ++(0, -0.4) -| (-3.8, -1.5);

\node[aer] (env) at (-3.8, -1.9) {envelope.json};
\node[aer] (pln) at (-3.8, -2.7) {plans.jsonl};
\node[aer] (stp) at (-3.8, -3.5) {steps.jsonl};
\node[aer] (vrd) at (-3.8, -4.3) {verdict.json};

\draw[arr] (agent) -- (0, -1.5);

\node[tool, minimum width=2.2cm] (mcp) at (0, -1.9) {MCP servers};
\node[tool, minimum width=2.4cm] (shell) at (0, -2.8) {Shell commands};
\node[tool, minimum width=2.2cm] (fio) at (0, -3.6) {File I/O};

\node[lbl] at (3.8, -0.8) {Ground truth};
\node[sublbl] at (3.8, -1.1) {Interceptors write};

\node[intercept] (imcp) at (3.8, -1.9) {Stdio interceptor};
\node[intercept] (ishell) at (3.8, -2.8) {Shell trap};
\node[intercept] (ifile) at (3.8, -3.6) {File watcher};

\draw[arr] (mcp.east) -- (imcp.west);
\draw[arr] (shell.east) -- (ishell.west);
\draw[arr] (fio.east) -- (ifile.west);

\node[sublbl] at (3.8, -4.2) {Interceptor logs (read-only to agent)};

\draw[darr] (-3.8, -4.7) -- (-3.8, -5.8) -- (-1.6, -6.2);
\draw[darr] (3.8, -4.5) -- (3.8, -5.8) -- (1.6, -6.2);

\node[sublbl] at (-3.8, -5.5) {AER files};
\node[sublbl] at (3.8, -5.5) {Interceptor logs};

\node[recon, minimum height=0.9cm] (rec) at (0, -6.5) {\textbf{Reconciler}\\\scriptsize Compares self-report vs ground truth};

\node[score, minimum width=3.2cm] (fid) at (0, -7.6) {\textbf{Fidelity score} $\in [0,1]$};
\draw[arr] (rec) -- (fid);

\end{tikzpicture}
\caption{AER dual-layer architecture. The agent produces self-reported AER files (left) while three independent interceptors---MCP stdio proxy, shell DEBUG trap, and filesystem watcher---capture ground-truth execution data (right). The reconciler compares both records post-session, producing a per-investigation fidelity score written to AER metadata by the verification process, not by the agent.}
\label{fig:architecture}
\end{figure*}

\textbf{Layer 1: MCP stdio interceptor.} In containerized agent deployments, MCP servers communicate via stdin/stdout JSON-RPC. At build time, each MCP server binary is wrapped with a transparent proxy that logs every JSON-RPC request (method, parameters) and response (result, error) to a separate JSONL file. The agent cannot distinguish the wrapper from the real binary, and the log directory is write-protected from the agent's user.

\textbf{Layer 2: Shell command interceptor.} Agent frameworks execute shell commands (\texttt{grep}, \texttt{curl}, \texttt{python3}) via bash subprocesses. A \texttt{BASH\_ENV}-triggered DEBUG trap logs every command invocation---binary name, arguments, timestamp, PID---before execution. The trap is set at container build time; the agent cannot disable it.

\textbf{Layer 3: Filesystem watcher.} A background process monitors file creation, modification, and deletion across the agent's working directories, logging events with timestamps and paths. This catches cases where the agent reports writing to one file but actually writes to another, or reports no file activity while silently creating artifacts.

All three logs are written to a directory the agent can read but cannot modify, and are volume-mounted to the host for post-session analysis.

\textbf{Reconciler.} After each session, a reconciliation script cross-references AER's \texttt{steps.jsonl} against the three interceptor logs, checking:

\begin{itemize}
\item \textbf{Coverage}: every interceptor-logged call has a corresponding AER step, and vice versa. Unmatched entries indicate hidden or fabricated actions.
\item \textbf{Tool name match}: the tool name recorded in AER matches the actual JSON-RPC method or command executed.
\item \textbf{Input match}: the parameters recorded in AER match the actual request payload. This is the check that detects the class of error illustrated by our running deployment---an agent reporting ``search in region Montreal'' while actually sending ``region=Ashburn'' in the tool call.
\end{itemize}

The reconciler produces a \textbf{fidelity score} $\in [0,1]$ written to the AER's \texttt{metadata.json}---not by the agent, but by the independent verification process. The score decomposes into MCP coverage, shell coverage, tool match rate, fabrication rate, and hidden call rate. This provides a per-investigation, independently-verified trust metric for AER data quality.

This architecture does not solve the faithfulness problem for self-reported \emph{reasoning} (intent, observation, inference remain the agent's own framing), but it establishes ground truth for the \emph{behavioral trace}: which tools were called, with what inputs, in what sequence. Mismatches between self-reported intent and verified behavior---the agent says it searched one region but verifiably searched another---are detectable from the reconciliation output. The behavioral trace exposes the error without requiring faithful reasoning.

\section{The Agent Execution Record Model}

We present the AER model using the DBINFRA-1458 running example to illustrate each component.

\subsection{File Structure}

\begin{lstlisting}
agent-executions/incidents/DBINFRA-1458/
  envelope.json  # Identity, trigger, delegation
  plans.jsonl    # Plan v1, Plan v2 (revised)
  steps.jsonl    # Steps 1-4 with reasoning
  verdict.json   # OOM Kill + evidence chain
  metadata.json  # Duration, cost, pin status
\end{lstlisting}

\subsection{Envelope: Identity, Authority, and Retrieved Context}

The envelope is written once at investigation start. For DBINFRA-1458:

\begin{lstlisting}
{"investigation_id": "DBINFRA-1458",
 "trigger": {"source":"jira-sd",
   "reference":"DBINFRA-1458",
   "summary":"DB connectivity failures
              on SCAN IP",
   "severity":"P2"},
 "agent": {"agent_version":"rca-v2.4.1",
   "model":"codex-5.3",
   "prompt_version":"rca-prompt-v7.2"},
 "authority": {
   "delegated_by":"oncall:sre-payments",
   "delegation_mechanism":
     "iam-policy:rca-agent-role-v3",
   "permissions_scope":
     ["monitoring:read:payments"],
   "authority_chain": [
     {"principal":"incident-auto-trigger",
      "type":"system"},
     {"principal":"sre-payments-team",
      "type":"team"},
     {"principal":"rca-agent-v2.4.1",
      "type":"agent"}]},
 "context_snapshot": {
   "retrieval_context": {
     "sources": [{"type":"rag",
       "query":"SCAN IP failure RAC",
       "chunks_retrieved":3,
       "chunk_ids":["rb-0042","rb-0108",
                    "inc-2891"],
       "total_tokens":2840}]},
   "system_context":
     {"service":"payments-db",
      "region":"us-ashburn-1"}}}
\end{lstlisting}

The \textbf{delegation chain} records under whose authority the agent acts. The \textbf{retrieval context} captures what actually entered the agent's context window (specific RAG chunk IDs, token counts) rather than what the agent could have queried---following a principle of \emph{retrieval provenance over availability provenance}.

\subsection{Versioned Plans}

For DBINFRA-1458, the agent formulates an initial plan and later revises it:

\begin{lstlisting}
{"plan_version":1,
 "rationale":"DB connectivity issue.
   Check: 1) SCAN DNS, 2) listeners,
   3) network, 4) recent changes.",
 "steps_intended":["check_dns",
   "check_listeners","check_net",
   "review_changes"]}
{"plan_version":2, "supersedes":1,
 "revision_trigger":"step_002",
 "rationale":"Listener down on node 3.
   Deep-dive: instance status, logs,
   memory/reboot events.",
 "steps_intended":["check_node3_crs",
   "check_node3_syslogs"]}
\end{lstlisting}

The plan version chain captures \emph{adaptive reasoning}---the agent's strategic response to unexpected evidence. The \textbf{revision\_trigger} pointer (\texttt{step\_002}) explicitly links the re-plan to the observation that caused it.

\subsection{Steps with Reasoning Provenance}

Each step captures the intent/observation/inference triple alongside mechanical tool calls. Steps 1 and 2 of DBINFRA-1458:

\begin{lstlisting}
{"step_id":"step_001","plan_version":1,
 "sequence":1,
 "intent":"Verify SCAN DNS resolution
   to rule out network-layer cause",
 "tool_calls":[{"tool":"nslookup",
   "input":"payments-scan.internal",
   "output":"Resolves to 10.0.1.50,.51,.52
     - all 3 VIPs",
   "duration_ms":120}],
 "observation":"SCAN DNS is healthy.
   All 3 VIPs responding.",
 "inference":"DNS ruled out.
   Move to listener verification.",
 "tokens":{"input":847,"output":213}}
{"step_id":"step_002","plan_version":1,
 "sequence":2,
 "intent":"Check RAC listener status on
   all nodes to identify failing node",
 "tool_calls":[{"tool":"lsnrctl_status",
   "input":"all_nodes",
   "output":"Node1:OK, Node2:OK,
     Node3:TNS-12541 no listener",
   "duration_ms":340}],
 "observation":"Node 3 listener is down.
   Directly explains connectivity failures.",
 "inference":"Proximate cause identified.
   Does not explain root cause.
   Re-planning to investigate Node 3.",
 "tokens":{"input":890,"output":313}}
\end{lstlisting}

Note the progression: step 1's inference (``rule out DNS, move to listeners'') leads to step 2's intent (``check listeners to identify failing node''). Step 2's inference triggers plan revision. This causal chain is captured as structured data, not as text buried in a model response.

\subsection{Evidence Chain and Structured Verdict}

After investigating Node 3's CRS status and system logs (steps 3--4), the agent produces:

\begin{lstlisting}
{"root_cause_category":
   "Infrastructure > Memory > OOM Kill",
 "root_cause_summary":"OOM on Node 3
   killed PMON, triggering watchdog reboot.
   CRS failed auto-restart, leaving listener
   and instance offline, causing SCAN
   connectivity failures.",
 "confidence":0.95,
 "affected_components":["node3",
   "payments-db-cluster","payments-service"],
 "evidence_chain":
   ["step_002","step_003","step_004"],
 "alternatives_rejected":[
   {"hypothesis":"SCAN DNS misconfiguration",
    "rejected_by":"step_001",
    "reason":"All 3 VIPs resolving"},
   {"hypothesis":"Network partition",
    "rejected_by":"step_001",
    "reason":"All nodes reachable"}],
 "remediation":["Start CRS on Node 3",
   "Investigate memory consumer",
   "Add OOM monitoring alert"]}
\end{lstlisting}

The \textbf{evidence chain} (steps 2$\rightarrow$3$\rightarrow$4) is distinct from the full execution tree (steps 1$\rightarrow$2$\rightarrow$3$\rightarrow$4). Step 1 was exploratory and did not contribute to the verdict. The \textbf{alternatives rejected} with disconfirming step pointers capture the agent's hypothesis elimination process.

\subsection{Domain Profiles}

The AER core is domain-agnostic. Domain profiles extend steps and verdicts with domain-specific vocabulary: RCA (root cause taxonomy, blast radius), Coding (files modified, test results), Security (MITRE ATT\&CK mapping), Deployment (canary metrics, rollback decisions). Profiles are versioned independently of the core.

\section{Replay: From Records to Regression Suites}

\subsection{Three Modes}

\textbf{Narrate:} Read-only step-by-step walkthrough. Zero cost, zero execution. For debugging, incident reviews, onboarding.

\textbf{Mock:} Re-runs reasoning with recorded tool outputs under a different model or prompt version. The agent receives exact same tool results but reasons with a new model. For each step, the report compares: did the new model reach the same observation? The same inference? At the end: did it reach the same verdict? This is, to our knowledge, not natively provided by any existing system. LangGraph's time-travel resumes against live systems. LangSmith's evaluation uses synthetic inputs. Mock replay uses \emph{real production investigations as test cases, automatically, with zero live API calls}.

\textbf{Live:} Re-executes the original plan against live systems. Different results (world has changed). Useful for verifying resolution.

\subsection{Mock Replay: Running Example}

Suppose we upgrade from codex-5.3 to codex-6.0 and mock-replay DBINFRA-1458:

\begin{lstlisting}
$ aer replay DBINFRA-1458 \
    --mode mock --model codex-6.0

Mock replay: DBINFRA-1458
  original: codex-5.3, new: codex-6.0

step_001: "DNS ruled out. Check listeners."
  new:     "DNS healthy. Listeners next."
  >> Functionally equivalent

step_002: "Proximate cause. Re-planning."
  new:     "Listener down. Check CRS."
  >> Divergence: new model skips re-plan

step_004: Verdict: OOM Kill (0.95)
  new:     Verdict: OOM Kill (0.91)
  >> Verdict converged

Summary: 3/4 matched. Verdict converged.
Note: codex-6.0 is more direct
      (skips explicit re-plan step)
\end{lstlisting}

Batch mock-replay across 50 pinned investigations produces a population-level behavioral comparison: verdict convergence rate, re-plan frequency shift, hypothesis breadth change.

\section{Architecture and Preliminary Storage Analysis}

\subsection{Local-First File-Based Capture}

The reference implementation writes JSONL files to a configurable directory. Zero infrastructure required. A Python SDK exposes: \texttt{start\_investigation()}, \texttt{log\_plan()}, \texttt{log\_step()}, \texttt{record\_verdict()}. Automatic lifecycle management: count-based eviction (default 50), time-based eviction (default 14 days), pin/promote workflow.

\subsection{Scale-Out Streaming Architecture}

For enterprise deployment: async emission to a Kafka-compatible streaming platform partitioned by investigation ID; materializer assembles complete AER documents; wire format transitions from JSONL to Avro for compactness and schema enforcement; hot storage in a JSON-native document database; search index for full-text exploration.

\subsection{Preliminary Storage Comparison}

\emph{The following analysis is based on a stylized 10-step investigation and should be treated as preliminary motivation, not a validated empirical result. Actual compression ratios will vary with workload characteristics.}

\begin{table}[H]
\centering
\small
\caption{Preliminary storage comparison for a 10-step investigation.}
\label{tab:storage}
\begin{tabular}{@{}p{2.6cm} p{2.0cm} p{2.0cm}@{}}
\toprule
\textbf{Metric} & \textbf{Cum.\ Chkpt} & \textbf{AER} \\
\midrule
Per investigation & $\sim$560 KB & $\sim$25--130 KB \\
Estimated ratio & 1$\times$ & 4--22$\times$ compact \\
Value trajectory & Decays & Appreciates \\
Retention model & Binary & Graduated \\
\bottomrule
\end{tabular}
\end{table}

\section{Evaluation Methodology}

\emph{This section describes our planned evaluation methodology. Preliminary deployment informs the design; full empirical results are ongoing work.}

\subsection{Behavioral Analytics Enablement}

We define 10 population-level behavioral analytics questions that a platform team operating autonomous agents needs to answer. For each, we assess whether it is answerable from (a) AER metadata in seconds, (b) checkpoint data with custom extraction, (c) observability trace data with custom pipelines. We hypothesize that AER provides immediate answers for all 10; checkpoints and traces require significant custom work for 2--4 and cannot practically address the remaining 6--8.

\subsection{Mock Replay Regression Testing}

Record AERs for 50 incidents with raw capture. Batch mock-replay substituting a different prompt version. Measure: verdict convergence rate, evidence chain overlap, reasoning pattern divergence.

\subsection{Storage Economics (Empirical)}

Measure actual storage for 100 real investigations represented as (a) LangGraph checkpoints, (b) AER without raw, (c) AER with raw, (d) AER with value-based compaction.

\subsection{Expressiveness Comparison}

For 20 sampled incidents: can the representation answer (a) ``why step 3?'' (intent), (b) ``why plan changed?'' (plan versioning), (c) ``which evidence?'' (evidence chain), (d) ``under whose authority?'' (delegation), (e) ``what context seen?'' (retrieval provenance). Score each representation.

\subsection{Faithfulness Validation}

We evaluate AER fidelity along two dimensions. \textbf{Behavioral fidelity} is assessed automatically via the reconciler (Section~3.4): for each investigation, the fidelity score measures agreement between self-reported tool calls and independently intercepted execution data. Across $N$ investigations, we report mean fidelity, distribution of mismatch types (fabricated, hidden, tool name divergence, input divergence), and correlation between fidelity score and investigation complexity (step count, plan revisions).

\textbf{Reasoning fidelity} is assessed via expert review: (a) expert rating of intent correctness on 50 randomly sampled steps, (b) action/intent consistency score (do agents that report ``rule out DNS'' actually perform DNS checks, as verified by the interceptor?), (c) predictive validity of confidence (does higher AER confidence predict higher expert agreement?). Note that (b) is now mechanically verifiable via the reconciler rather than requiring manual inspection---the interceptor log confirms whether a DNS-related tool was actually invoked.

\section{Discussion}

\subsection{Complementarity}

AER occupies a different analytical level from state checkpoints and observability platforms. The relationship is layered: checkpointers handle operational resilience; observability platforms handle per-run debugging; audit trails handle compliance; AER handles population-level behavioral analytics. A production deployment benefits from all four. AER does not replace existing systems; it fills an analytical gap that emerges at scale and autonomy.

\subsection{Adoption Cost and Agent Cooperation}

Unlike passive checkpoints and traces, AER requires the agent to produce structured reasoning annotations. This means prompts must elicit intent/observation/inference as structured output, and the SDK must be integrated into the execution loop. We consider this a lightweight adoption cost and note that requesting structured reasoning annotations often improves response quality by imposing organizational discipline on the model's output. However, the quality of AER data depends on prompt engineering quality.

\subsection{Schema Evolution and Privacy}

Core schema uses additive-only evolution (new fields, never removing old ones). Domain profiles versioned independently. Raw capture is configurable per-deployment with field-level redaction for promoted AERs. The delegation chain is subject to IAM policy visibility constraints.

\subsection{Beyond Single-Agent Records}

Multi-agent investigations require a higher-order Investigation Record composing multiple AERs from collaborating agents. The deterministic investigation ID provides the join key. Cross-agent evidence chain comparison is a natural application of AER's structured verdict. We consider this important future work.

\section{Conclusion}

Existing systems provide strong operational tooling for individual agent executions: state checkpoints for fault tolerance, execution traces for debugging, telemetry for performance monitoring, audit trails for compliance. What they do not natively provide as a first-class, schema-level construct is structured reasoning provenance: normalized, queryable records of why the agent made each decision, how observations shaped its strategy, which evidence supports its conclusions, and how its reasoning patterns behave across populations.

Agent Execution Records address this gap by capturing the reasoning layer as first-class structured data at execution time, enabling population-level behavioral analytics---reasoning pattern mining, confidence calibration, cross-agent comparison, and counterfactual regression testing via mock replay---that is difficult or impractical to achieve through post-hoc extraction from existing artifacts. The transport-layer verification mechanism (Section~3.4) addresses the self-report trust problem by providing independently captured ground truth against which AER's behavioral trace can be reconciled, producing per-investigation fidelity scores without requiring access to the model's internal computation.

AER complements existing infrastructure rather than competing with it, occupying a different analytical layer in the agent stack. As organizations transition from agent experimentation to platformized autonomous agents at enterprise scale, the ability to systematically understand, test, and improve agent reasoning becomes essential infrastructure. Agent Execution Records provide a foundation for that capability.


\end{document}